# LAMBDA: Covering the Solution Set of Black-Box Inequality by Search Space Quantization

Lihao Liu[1], Tianyue Feng, Xingyu Xing, Junyi Chen[2]

School of Automotive Studies, Tongji University
leoherz_liu@163.com[1], chenjunyi@tongji.edu.cn[2]

## Abstract

Black-box functions are broadly used to model complex problems that provide no explicit information but the input and output. Despite existing studies of black-box function optimization, the solution set satisfying an inequality with a black-box function plays a more significant role than only one optimum in many practical situations. Covering as much as possible of the solution set through limited evaluations to the black-box objective function is defined as the Black-Box Coverage (BBC) problem in this paper. We formalized this problem in a sample-based search paradigm and constructed a coverage criterion with Confusion Matrix Analysis. Further, we propose LAMBDA (Latent-Action Monte-Carlo Beam Search with Density Adaption) to solve BBC problems. LAMBDA can focus around the solution set quickly by recursively partitioning the search space into accepted and rejected subspaces. Compared with La-MCTS, LAMBDA introduces density information to overcome the sampling bias of optimization and obtain more exploration. Benchmarking shows, LAMBDA achieved state-of-the-art performance among all baselines and was at most 33x faster to get 95% coverage than Random Search. Experiments also demonstrate that LAMBDA has a promising future in the verification of autonomous systems in virtual tests.

## 1 Introduction

Black-box functions refer to functions with no explicit structures and information (e.g., gradient) but the input and output for utilization. Black-box functions are usually faced with a high cost of time and economy. Thus, a limited number of times to evaluate the objective function can be given.

Optimization of black-box functions has been widely studied. Many Black-Box Optimization (BBO) algorithms, such as Genetic Algorithm (GA), Differential Evolution (DE) and Bayesian Optimization (BO) are proposed to solve the BBO problem by seeking for the global optimum according to the input and output of the black-box objective function. The BBO problems can be denoted as Eq. (1), in which $x$ means a point in search space, $f$ means the objective function.

$$x^* = arg \max_x f(x) \quad (1)$$

However, in some practical issues, there is a need to cover as comprehensive a range of feasible solutions as possible, rather than one optimum. The feasible solutions can be defined as solutions satisfying Eq. (2), in which $\delta$ means the criterion of the feasibility of solutions.

$$f(x) > \delta \quad (2)$$

The Black-Box Coverage (BBC) problem emerges, how to identify the solution set of a black-box inequality with limited evaluation opportunities to the black-box function?

Inspired by La-MCTS (Wang, Fonseca, and Tian 2020), we developed an algorithm named LAMBDA (Latent-Action Monte-Carlo Beam Search with Density Adaption) to solve BBC problems. LAMBDA can quickly focus on the possible solution sub-space by recursively partitioning the search space into good and bad regions and dynamically adjust the hierarchy partition towards more accuracy with a UCB (Upper Confidence Bound) bandit algorithm. The main differences of LAMBDA compared with La-MCTS include: A) an adaptive kernel density estimator (KDE) to overcome the sampling bias in La-MCTS, which would lead to over-exploitation during optimization. Thus, the learning of search space partition and UCB calculation was redesigned. B) the Beam Search technique was adapted to improve the parallelizability and enhance the exploration.

The main contributions of this paper are: A) proposition and formalization of the BBC problem, B) the LAMBDA algorithm for solving BBC problems, and C) experiments to verify the effectiveness of the proposed algorithm. The contents of this paper are organized as below: Section 1, introduction. Section 2, related works. Section 3, the formalization of the BBC problem. Section 4, LAMBDA algorithm. Section 5, benchmarking and experiments. Section 6, discussion. And Section 7, conclusion.

## 2 Related Works

The most relevant topic of our work is Black-Box Optimization algorithms, which can be divided into 3 categories: population-based, surrogate-based, and MCTS-based.

Population-based methods are inspired mainly by the behavior of the biological population. Genetic Algorithm (Henry, 1992) simulates evolution and selection, uses cross and mutation operators to propose new samples. Differential Evolution (Storn and Price 1997) is similar to GA but uses vector differences to perturbate current samples. Besides, Particle Swarm Optimization (Kennedy and Eberhart 1995), CMA-ES (Hansen, Müller, and Koumoutsakos 2003), Simulated Annealing (Kirkpatrick, Gelatt Jr, and Vecchi 1983) are also broadly known in the population-based optimization methods. We chose the most classic GA and DE as baselines in benchmarking from the population-based methods.

Surrogate-based methods maintain a surrogate model of the objective function during optimization, and determine candidates to be evaluated obeying to the surrogate. Bayesian Optimization (Pelikan, Goldberg, and Cantú-Paz 1999) is a typical surrogate-based method succussing in many fields such as NAS (White, Neiswanger, and Savani 2021), structure design (Yamawaki et al. 2018), and hyperparameter tunning (Snoek, Larochelle, and Adams 2012). The most significant limitation of BO is the $O(n^3)$ complexity of the Gaussian Process Regressor surrogate, leading to inefficacy on problems over 10 dimensions and several thousand evaluations. TPE (Bergstra et al. 2011) and BOHB (Falkner, Klein, and Hutter 2018) replace the GPR surrogate with Parzen Estimator to obtain higher performance. While TuRBO (Eriksson et al. 2019) introduces trust region, restarting, and implicit multi-armed bandit to adapt BO to large scale. We compared LAMBDA with both BO and TuRBO in this paper.

MCTS-based methods are most relevant to our work. Despite its fantastic performance in Go (Silver et al. 2016), MCTS can be used to solve optimization problems. Search space partition is a common technique to adapt MCTS to continuous problems in this line of research. DOO (Munos 2011), SOO (Munos, 2011), and HOO (Bubeck et al. 2011) use k-ary partitions, while Kim et al. (2020) introduce efficient Voronoi partitions. Recently, Wang et al. proposed LA-NAS (2020) and LA-MCTS (2020) to learn arbitrary decision boundaries of partitions. Wang et al. (2021) show linear partitions performing better than curving partitions on the whole. LAMBDA inherits MCTS's framework and the thought of search space partition from the above study but introduces density information to overcome the sampling bias of optimization. For MCTS-based methods, we take the most state-of-the-art La-MCTS into benchmarking.

## 3 Definition of the BBC Problem

The BBC problem can be defined as follow. Table.1 shows the denotations in this paper. Given a black-box function $f$ and a criterion $\delta$, the goal is to figure out the solution set satisfying the inequality $f(x) > \delta$. In this paper, only deterministic functions are considered. The question is that no information else except the input and output of $f$ can be obtained, and the number of times to evaluate $f$ is limited. It's a typical sample-based optimization problem that might be solved within an optimization paradigm.

The optimization agent $\Pi$ has a limited optimization budget $N$, which means the agent has only $N$ opportunities to sample from $f$. At each sampling from $f$, the agent decides where to place the sampling point $x$ according to the historical sample records $\mathcal{D}$, get an output $y$ through evaluating $f(x)$, and appends the new pair $(x, y)$ to $\mathcal{D}$.

$$(x, y) := (\Pi(\mathcal{D}), f(x))$$
$$\mathcal{D} := \mathcal{D} \cup (x, y) \quad (3)$$

After running out the optimization budget, the agent $\Pi$ gets sample records $\mathcal{D}$ of size $N$. A classifier $\mathcal{C}$ to predict if $x$ satisfying $f(x) > \delta$ can be learned from $\mathcal{D}$, which divides the search space $\Omega$ into sub-space $\widehat{\Omega}_\delta$ and sub-space $\widehat{\Omega}_{\neg\delta}$. $\widehat{\Omega}_\delta$ is the sub-space predicted satisfying $f(x) > \delta$ by $\mathcal{C}$, while $\widehat{\Omega}_{\neg\delta}$ is $\Omega - \widehat{\Omega}_\delta$.

With the actual $\Omega_\delta$ and $\Omega_{\neg\delta}$ known, the quality of $\mathcal{D}$ and performance of $\Pi$ and $\mathcal{C}$ can be assessed with Confusion Matrix Analysis (Powers 2011), as shown in Fig.1. In the Confusion Matrix, $recall$ represents the coverage of $\mathcal{C}$'s prediction to the solution set of $f(x) > \delta$. However, an intentional classifier can decide all points in the search space be satisfying $f(x) > \delta$, then $recall$ would be 100%. For all practical purposes it makes no sense. The $precision$ must be taken into consideration. In this paper, the $F_2$ score is used to balance $recall$ and $precision$, with $recall$ being more emphasized.

|  | $\Omega_\delta$ | $\Omega_{\neg\delta}$ |  |
|---|---|---|---|
| $\widehat{\Omega}_\delta$ | TP | FP | $precision$<br>$precision = \frac{TP}{TP + FP}$ |
| $\widehat{\Omega}_{\neg\delta}$ | FN | TN |  |
| | $recall$<br>$recall = \frac{TP}{TP + FN}$ | | $F_2$<br>$F_2 = \frac{5 \cdot (recall + precision)}{4 \cdot precision + recall}$ |

Fig.1 The Confusion Matrix of BBC problem

The key point of the BBC problem is how to acquire information about the search space with a limited optimization budget. This paper focuses on the optimization agent $\Pi$, which decides where to sample from the search space and acquire information about $f$. For the classifier $\mathcal{C}$, appropriate methods can be chosen according to domain knowledge of each practical issue. However, for generality and fairness, a basic configuration of $\mathcal{C}$ was used in all experiments of this paper. A regressor $\hat{f}$ of $f$ was built from $\mathcal{D}$ with *SciPy*'s linear interpolation method (SciPy, 2021). Thereby the classifier can be constructed with $\hat{f}$ simply.

$$\mathcal{C}(x) = 1 \text{ IF } \hat{f}(x) > \delta \text{ ELSE } 0 \quad (4)$$

Table.1 Denotations.

| | | | |
|---|---|---|---|
| $\Omega$ | whole search space | $\Omega_i$ | sub-space of node $i$ |
| $f$ | objective function | $\hat{f}$ | regressor of the objective function |
| $(x, y)$ | the pair of a sample point and its result | $\delta$ | criterion in the inequality $f(x) > \delta$ |
| $\Omega_\delta$ | sub-space satisfying $f(x) > \delta$ | $\Omega_{\neg\delta}$ | sub-space not satisfying $f(x) > \delta$ |
| $\hat{\Omega}_\delta$ | predicted sub-space satisfying $f(x) > \delta$ | $\hat{\Omega}_{\neg\delta}$ | predicted sub-space not satisfying $f(x) > \delta$ |
| $\mathcal{D}$ | set of historical sample records | $\mathcal{D}_i$ | set of historical sample records located in $\Omega_i$ |
| $X$ | set of historical sample points | $X_i$ | set of historical sample points located in $\Omega_i$ |
| $Y$ | set of historical sample results | $Y_i$ | set of historical sample results located in $\Omega_i$ |
| $N$ | optimization budget | $n_i$ | number of samples in $\Omega_i$ |
| $v_i$ | average value of $\Omega_i$ | $\rho_i$ | average sampling density of $\Omega_i$ |
| $c_p$ | exploration factor | $w_i(x)$ | normalized weight of sample point $x$ in $\Omega_i$ |

## 4 LAMBDA Algorithm

### 4.1 Sampling Bias of La-MCTS

MCTS is an algorithm for solving discrete sequential decision problems, widely known for its surprising performance in Go (Silver et al. 2016). However, MCTS is not originally applicable to continuous optimization because it requires a fixed and limited action space, which is usually arbitrary and infinite in continuous optimization problems. La-MCTS breaks this limitation by learning partitions of search space to form latent actions and adapts MCTS to optimization problems in continuous search space.

However, La-MCTS is faced with a further question called sampling bias. As the optimization proceeds, samples concentrate towards more hopeful sub-spaces. Hence the records set suffers from an imbalance of samples, leading to that: A) the partition learned from $\mathcal{D}$ over-emphasizes these hopeful regions and cuts them into many fragmental sub-spaces, B) the UCB score based on the mean value and number of samples in a sub-space loses efficacy. The two impacts above make the optimization agent trapped in one local modal of the search space. Fig.2 shows an intuitive example of this phenomenon.

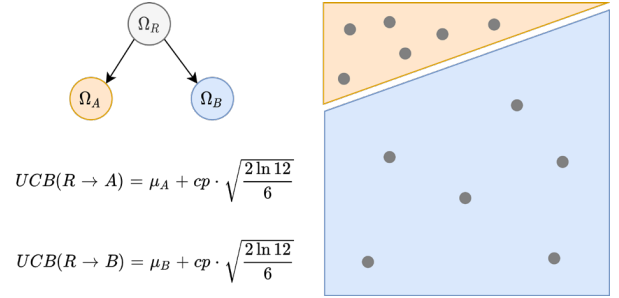

$$UCB(R \to A) = \mu_A + cp \cdot \sqrt{\frac{2\ln 12}{6}}$$

$$UCB(R \to B) = \mu_B + cp \cdot \sqrt{\frac{2\ln 12}{6}}$$

Fig.2 An Intuitive Example of Sampling Bias. $\Omega_A$ is the hopeful region of the search space. Compared with $\Omega_B$, it has a higher mean value of samples, a smaller volume, and the same number of samples. The exploration terms constructed with the numbers of samples turn out to be equal, and no longer encourage sampling in the under-explored sub-space $\Omega_B$.

Density information is introduced to overcome the sampling bias. The Inverse Probability Weighting (IPW) technique is used to re-balance the sample records in search space partitioning and UCB calculation, which will be detailed later.

In order to estimate the sampling density efficiently, we developed an adaptive Kernel Density Estimator (KDE) with Approximate Nearest Neighborhood (ANN). The bandwidth of the kernel is dynamically calculated by querying the $k$-th nearest neighbors' minimal bounding sphere so that the KDE can adapt to different problems without tunning hyperparameters. Besides, *faiss* (Johnson et al., 2019) was utilized to improve retrieval efficiency. The speedup structure (or *Index* in *faiss*) need to be re-trained periodically, because ANN assuming data is drawn from one fixed distribution while the distribution of samples gradually concentrates to target sub-spaces during the optimization.

### 4.2 LAMBDA

The framework of LAMBDA is similar with MCTS, consisting of extension (treeification), selection, simulation, backpropagation etc., as shown in Fig.3.

- Initial Sampling. Sample an initial $\mathcal{D}$ to construct the partition tree. Sobol sequence was used in sampling to get more uniformity.
- Treeification. Treeify the search space into sub-spaces recursively to form a quantization of the search space. Each leaf node in the partition tree represents a sub-space. We used linear partitions in this paper. More complex forms of partitions can be used, but with more calculation burden in the treeification and simulation stages. Sampling density estimated by KDE is introduced into this stage to get unbiased partitions.
- Selection. Select a batch of sub-spaces to sample according to the UCB scores of nodes in the partition tree. A new UCB calculation method is proposed based on the sampling density to overcome sampling bias.
- Simulation. Sample from the selected sub-spaces. LAMBDA just picks a random point in a sub-space with Rejected Sampling. The efficiency of Rejected Sampling can be accepted because of the linear partitions.
- Back-Propagation. Back-propagate new samples to update the partition tree. The partition tree is re-constructed with a period. Back-propagation is needed within one period.

Fig.3 The Structure of LAMBDA

### 4.2.1 Latent Action
A latent action is the partition of a search space or sub-space learned from historical samples, forms a decision boundary to split the parent space into two children, the good and the bad. The good one holds samples with higher objective function values. By recursively splitting the search space, a tree structure of partition is built on the search space. Then LAMBDA could utilize the partition tree to decide to which sub-spaces should be paid attention. Moreover, stop conditions are required to control the recursive partition process. We used $leafsize$ and $depth$ as stop conditions: A) if samples in a sub-space are less than $leafsize$, the partition on this sub-space will be stopped. B) if the branch depth of a leaf node reaches $depth$, the partition on this node will be stopped.

LAMBDA uses a KNN-SVM method to learn latent actions like La-MCTS (Wang, Fonseca, and Tian 2020), which firstly clusters the samples into two groups to get pseudo labels, then uses pseudo labels to train an SVM classifier to get the decision boundary. However, LAMBDA weights the samples by the normalized inverse of each sample's density to overcome the sampling bias. Samples in sparse areas get more importance than in dense areas. Fig.4 shows a comparison of partition between before and after weighting.

Fig.4 The Comparison of Partition between Before and After Weighting. The before one over-partitions the sub-space at the top-right corner and forms many fragments.

### 4.2.2 Multi-Beam Search
Beam Search is a technique broadly used in searching on graphs. Both the branch of best value and the other top-$k$ branches will be evaluated in Beam Search. It can improve exploration from pure greedy search and parallelizability by $k$ times with evaluating $k$ branches concurrently.

LAMBDA adapts beam search to MCTS by flatting the partition tree, as shown in Fig.5. There are only root and leaf nodes in the flatten partition tree. Therefore, when calculating UCB scores to do selection, A) all parent nodes in UCB expression should be the root node, B) top-$k$ branches of highest UCB scores should be selected all at once.

Fig.5 Flatting Partition Tree to Do Beam Search. Before flatting 3 selections and backpropagations are needed,

while after flatting the top-3 branches can be selected and simulated at once.

### 4.2.3 Density Adaptive

A new UCB calculation method was designed to replace the ineffective $UCB_1$(Eq. (5)) based on mean value and number of visits.

$$UCB_1(A \to B) = \frac{\sum_{x \in \mathcal{D}_B} f(x)}{n_B} + 2 \cdot c_p \cdot \sqrt{\frac{2 \ln n_A}{n_B}} \quad (5)$$

The new UCB score is based on density, and we named it $UCB_\rho$. $UCB_\rho$ brings sampling density into consideration to re-balance the records set $\mathcal{D}$ and get an unbiased estimation of the exploitation term and the exploration term. Eq. (6) describes the new UCB score.

$$UCB_\rho(A \to B) = \sum_{x \in \mathcal{D}_B} f(x) w_B(x) + c_p \cdot \ln\left(\frac{\rho_A}{\rho_B}\right) \quad (6)$$

In which,

$$w_B(x) = \frac{1/\rho(x)}{\sum_{x \in \mathcal{D}_B} 1/\rho(x)}$$
$$\rho_A = \sum_{x \in \mathcal{D}_A} w_A(x)\rho(x), \quad \rho_B = \sum_{x \in \mathcal{D}_B} w_B(x)\rho(x) \quad (7)$$

Here the mean sampling density of a sub-space can be estimated without extra sampling by utilizing in-hand samples from $\mathcal{D}$.

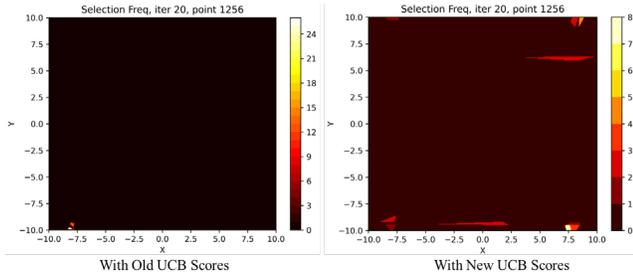

Fig.6 Sampling Frequency with Old and New UCB Scores. Samples over concentrated on one tiny point with the old UCB score, while more appropriately distributed with the new UCB score.

## 5 Experiments

We experimented with LAMBDA on an artificial function and a practical issue. In benchmarking, we chose baseline algorithms from the population-based, the surrogate-based, the MCTS-based and the basic random search, including GA, DE, BO, TuRBO, La-MCTS, RS (Random Search), and Sobol (Sobolev quasi-Monte-Carlo). Hyperparameters of baselines were set according to the suggestion of their authors. If no suggestions were found, these hyperparameters were set similar to those of LAMBDA, to ensure the fairness of benchmarking. To avoid randomness, each experiment was repeated 10 times, and each algorithm's mean performance was evaluated. Furthermore, the number of times to evaluate the objective function was used to assess the efficiency of algorithms, without bringing in confounders from platforms (e.g., OS) or hardware.

It's worth noting that BO (with GP) suffers from the complexity of $O(n^3)$. In our experiments, BO needs more than 60s to suggest the next batch of sampling points when the sample records exceed 2000. It's hard to accept the complexity of BO when the number of sample records grows large.

### 5.1 Benchmarking on Holder-Table Function

Holder-Table is a classical artificial function to verify performance of optimization algorithms. As shown in Fig.7, it has 4 global optimums distributed at the corners of the search space, while many local minima to trap the optimizer in local modals. The black-box inequality was defined as $f(x) > 18$, whose solution set was lined out with red color.

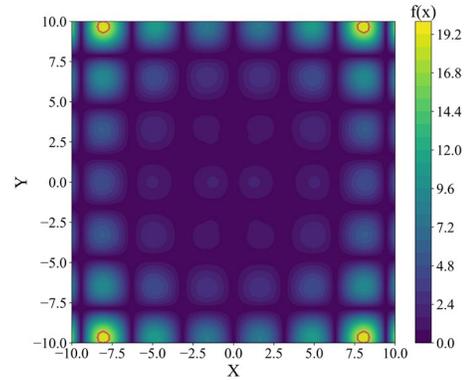

Fig.7 Holder-Table Function

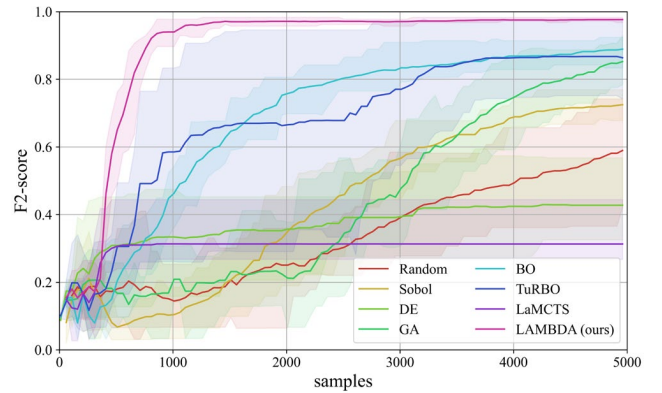

Fig.8 Benchmarking on Holder-Table Function

Benchmark results are shown in Fig.8. LAMBDA achieved the best performance among all algorithms in both mean F2 score, rate of convergence, and stability. TuRBO performed best among all baselines. In the best one of the ten trails, TuRBO obtained a similar F2 score and convergence rate with the worst performance of LAMBDA. But the large variance dragged down TuRBO's mean performance. La-MCTS didn't work well on this problem. It was trapped in one of the four modals due to the sampling bias. Thus, the F2 score always converged to about 25%. Besides, Sobol performed better than RS, indicating that sampling uniformity plays an important role in getting more coverage. The result of sampling dynamics shown in Fig.9 supports the analysis above as well.

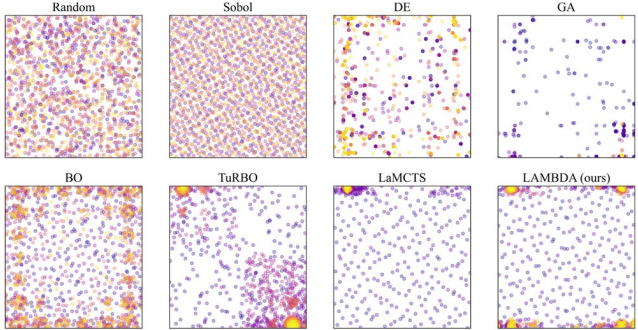

Fig.9 Sampling Dynamics of Each Algorithm at Iteration 1500. Color from purple to golden represents the record being sampled at the beginning or the end of the optimization.

### 5.2 Benchmarking on Safety Verification of Autonomous Driving System

Safety verification of an autonomous driving system in a particular logical scenario (Menzel, Bagschik, and Maurer 2018) is a typical BBC problem. We also benchmarked on this problem with a SIL (Software in Loop) simulation platform based on Virtual Test Drive (HEXAGON & MSC Software 2021).

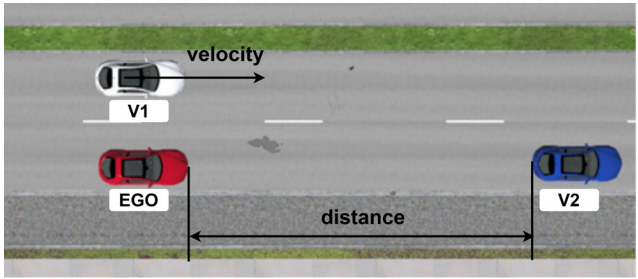

Fig.10 The Logical Scenario of Simulation

The simulation configuration is shown in Fig.10. The ego vehicle (EGO) is driven by an autonomous driving system (decision-making, planning, and control, with ideal perception), while surrounding vehicles V1 and V2 run according to parameter-defined movements. In this simulation, the parameters include the *distance* between EGO and V2, and the *velocity* of V1. Therefore, multiple modals of near-crash exist in the search space spanned by *distance* and *velocity*. The criterion TTC (Vogel 2003) is used to represent how close is the ego vehicle to collision. The scenarios with minimum TTC during simulation smaller than 0.5s are regarded critical, in which a crash is mostly unavoidable.

A fine-grained grid search was performed to get the ground truth of the objective function in the search space for benchmarking. Fig.11 shows the ground truth, red color lines out the critical sub-spaces.

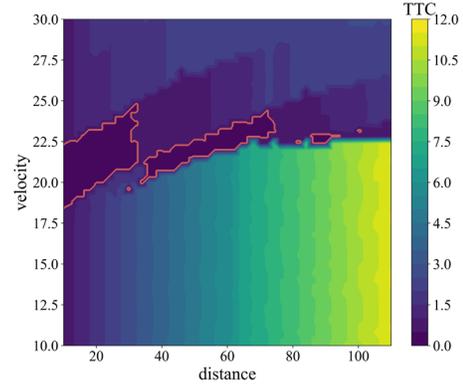

Fig.11 The Ground Truth of the Logical Scenario

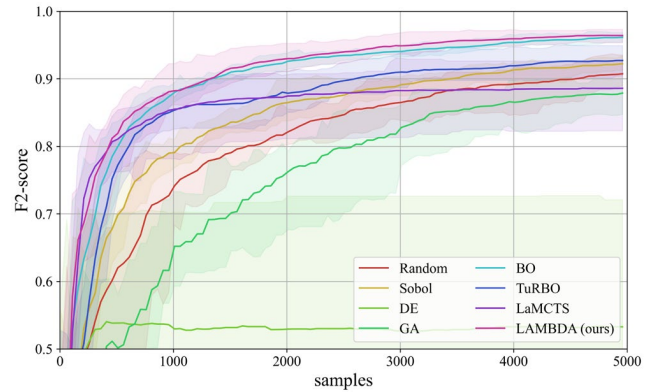

Fig.12 Benchmarking on the Logical Scenario

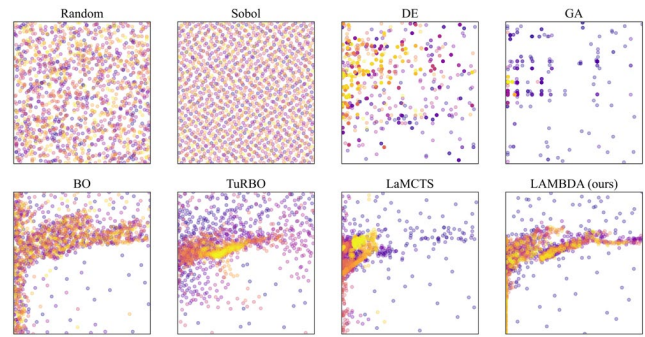

Fig.13 Sampling Dynamics of Each Algorithm at Iteration 1500 on the Logical Scenario

The benchmarking results are shown in Fig.12, while the sampling dynamics are shown in Fig.13. LAMBDA and BO

performed the best with LAMBDA gaining a slight advantage over BO. La-MCTS raised fastest among all algorithms initially but decelerated as optimization running, and finally converged to under 90%. F2 scores of all algorithms raised slowly when beyond 90%. The irregular distribution of critical sub-space makes it hard to cover the boundary completely.

## 6 Discussion

### 6.1 Tradeoff between Optimization and Coverage

Non-Free-Lunch Theory tells us no algorithm performs well on all optimization problems (Wolpert and Macready 1997). Tradeoff between optimization and coverage is needed. If an agent focuses more on coverage, it should explore the search space more to avoid missing important modals. However, more exploration means more budgets are spent riskily. Agents emphasizing optimization will quickly concentrate on a small search space region and dive into it to get as high objective function value as possible. It's why many algorithms like GA and DE perform well on BBO problems but might fail at BBC problems.

### 6.2 Comparation with BO

BO is more like a thought rather than an algorithm. The critical point is to use experience (or historical information) to predict the future and use new information to update experience. LAMBDA also follows this way. BO requires two kinds of information: A) expectation of the objective function and B) uncertainty of present estimation. In BO, GP regressors are usually used to model the above two kinds of information at every search space location. However, fitting the GP regressor is of $O(n^3)$ complexity.

To overcome the $O(n^3)$ complexity of BO, LAMBDA builds a partition tree on the search space, which forms quantization for the space, then only needs to learn rough regression to estimate the averaged expectation of subspaces. Density information represents the uncertainty of estimation, which can be retrieved efficiently from historical records with the ANN-accelerated KDE. Besides, the randomness of the algorithm makes quantization of the search space updated frequently, preventing the negative impact of roughening on the performance of LAMBDA.

## 7 Conclusion

In this paper, we proposed the Black-Box Coverage problem for practical purposes, which means to cover the solution set satisfying an inequality containing a black-box function. To achieve as high coverage as possible within a limited optimization budget, we developed the LAMBDA algorithm based on the idea of search space quantization. LAMBDA utilizes density information to overcome sampling bias and introduces Beam Search to obtain more parallelizability. Experiments demonstrate the state-of-the-art performance of LAMBDA on Black-Box Coverage problems. According to the benchmarking on Holder-Table function, LAMBDA can be 33x faster to get 95% coverage than Random Search. In future work, we plan to adapt LAMBDA to non-deterministic functions and look into high-dimensional search spaces.